\documentclass{article}
\usepackage{spconf,amsmath,graphicx}

\usepackage[table,xcdraw]{xcolor}
\usepackage[colorlinks]{hyperref}

\usepackage{graphicx}
\usepackage{multirow}
\usepackage{booktabs}
\usepackage{color, colortbl}
\usepackage{footmisc}
\usepackage{subfigure}
\usepackage{float}

\usepackage{enumitem}
\setlist{nosep, leftmargin=14pt}

\usepackage{mwe} % to get dummy images

\title{Medical semantic segmentation with diffusion pretrain}
\address{IRA Labs Ltd}
\name{David Li \qquad Anvar Kurmukov \qquad Mikhail Goncharov \qquad Roman Sokolov \qquad Mikhail Belyaev}
\begin{document}
\maketitle

% TODO
% Outperforming a closest non-contrastive method by 7.5% and performing on par with the state of the art contrastive method
\begin{abstract}
% In the past, the success of transformer architectures in pretext tasks provided a significant boost in the accuracy of solving various tasks. After that, there were multiple and quite successful attempts to transfer this property to models with 2D images. However, at the present time, in tasks involving 3D images, particularly medical images, it has not been possible to solve this task effectively. At the same time, recently popular diffusion models have demonstrated their ability to capture the semantics of images from datasets. Based on this, we attempted to use this model and adapt it to the domain of 3D data for pretraining on 3D images.

Recent advances in deep learning have shown that learning robust feature representations is critical for the success of many computer vision tasks, including medical image segmentation. In particular, both transformer and convolutional-based architectures have benefit from leveraging pretext tasks for pretraining. However, the adoption of pretext tasks in 3D medical imaging has been less explored and remains a challenge, especially in the context of learning generalizable feature representations.

We propose a novel pretraining strategy using diffusion models with anatomical guidance, tailored to the intricacies of 3D medical image data. We introduce an auxiliary diffusion process to pretrain a model that produce generalizable feature representations, useful for a variety of downstream segmentation tasks. We employ an additional model that predicts 3D universal body-part coordinates, providing guidance during the diffusion process and improving spatial awareness in generated representations. This approach not only aids in resolving localization inaccuracies but also enriches the model's ability to understand complex anatomical structures. 

Empirical validation on a 13-class organ segmentation task demonstrate the effectiveness of our pretraining technique. It surpasses existing restorative pretraining methods in 3D medical image segmentation by $7.5\%$, and is competitive with the state-of-the-art contrastive pretraining approach, achieving an average Dice coefficient of 67.8 in a non-linear evaluation scenario.

% We demonstrate the effectiveness of trained feature representation on a 10-class organ segmentation dataset, outperforming established non-contrastive learning methods for 3D medical image segmentation. Our method also performs on-par with the state-of-the-art contrastive pretraining approaches, achieving a 0.6 Dice score on a non-linear evaluation scenario. 

% The results indicate that our diffusion-based pretraining strategy is a significant step toward versatile and accurate 3D medical image segmentation, promising to enhance various clinical diagnostic procedures.

\end{abstract}
%
% \begin{keywords}
% Pretrain, Diffusion models, Medical images
% \end{keywords}
%

\section{Introduction}
\label{sec:intro}

Medical image segmentation has traditionally relied on supervised learning paradigms, which, despite their efficacy, come with notable drawbacks. It incurs high annotation costs but also leads to models that often suffer from poor generalization to new data sources. Representation learning via self-supervision (SSL) has emerged as a robust alternative, circumventing the need for labelled data by pretraining a model to extract meaningful representations from unlabeled data. This paradigm allows for the subsequent application of a linear or non-linear head on top of the pretrained backbone, which can be fine-tuned for various downstream tasks \cite{he2022masked}. Such a strategy is particularly advantageous in medical imaging, given the ever-expanding corpus of data on one hand, and on the other, a vast array of tasks for which the volume of labelled data is very small.

While contrastive learning has become a cornerstone of SSL in computer vision, its implementation in medical imaging has been met with mixed success. Existing methods that leverage contrastive learning for dense representations often fall short in resolving fine-grained voxel-level details \cite{taleb20203d}, rendering them suboptimal for high-resolution segmentation tasks. Recent dense SSL approaches \cite{haghighi2021transferable,tang2022self} have attempted to address these issues by combining restorative (generative, non-contrastive) learning with patch-level contrastive learning. However, a critical evaluation of these models in linear or non-linear probing setups, widely recognized as standard benchmark in the SSL for natural images, remains under-researched. Recently, this gap was bridged by authors of \cite{goncharov2023vox2vec}, who specifically assessed the \textbf{effectiveness of representations} learned through contrastive learning for medical image segmentation tasks.

We propose a non-contrastive,  diffusion model-based method of pretraining that is specifically tailored for 3D medical imaging. Unlike prior methods, our model is trained to generate 3D medical images through a process informed by anatomical guidance, thereby capturing the nuanced semantics of medical data more effectively. The main motivation is to generate locally informative representations, while introducing global information into patch-based training, via conditioning on anatomical coordinates.  Our main contributions are \textbf{two-fold}:
\begin{enumerate}
    \item We propose an anatomically guided diffusion-based pretraining procedure designed to produce informative feature representations for 3D images sourced from computed tomography.
    \item Through rigorous evaluation we demonstrate that informative voxel-level feature representations can be achieved not only through contrastive procedures, but also via non-contrastive approaches, resulting in a Dice coefficient of 67.8 in a non-linear probing scenario on a organs' segmentation task.
\end{enumerate}

\section{Related work}
\label{sec:format}
\begin{figure*}
\begin{center}
\includegraphics[scale=0.52]{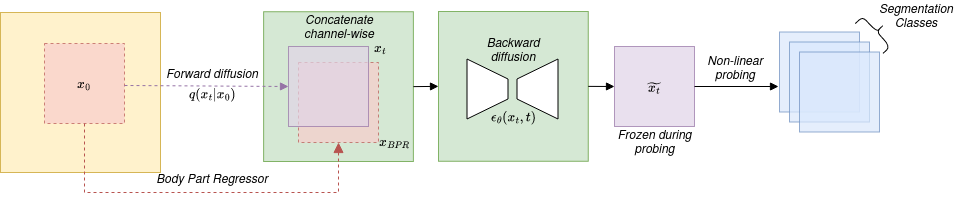}
\end{center}
   \caption{Model schema. Image patch $x_0$ is passed through diffusion process resulting in noised patch $x_t$. Independently, it is passed through a Body Part Regressor network to produce dense coordinate map $x_{BPR}$ of the same spatial resolution. $x_t$ and $x_{BPR}$ are then concatenated and passed through a trainable backward diffusion network to produce dense voxel-wise representations $\widetilde{x_t}$. For downstream supervised segmentation task a non-linear probing model is trained on top of frozen feature representations. All patches are \underline{three-dimensional}.}
\label{fig:pipeline}
\end{figure*}

\subsection{Self-supervised pretraining}
Self-supervised methods in medical imaging have evolved from basic pretext tasks, such as rotation and patch position prediction \cite{taleb20203d}, to advanced methods like restorative autoencoders \cite{haghighi2021transferable,tang2022self}, patch-level and voxel-level contrastive embedding techniques \cite{tang2022self,goncharov2023vox2vec}.

Contrastive methods rely on juxtaposing different parts of the same (or different) images, or even fine-grained image components such as voxels, to create an auxiliary learning task. For example, \cite{goncharov2023vox2vec} utilizes a Feature Pyramid Network (FPN) to generate similar voxel representations for the same voxels, obtained from different views of an image.

Restorative methods focus on learning to restore an image or an image patch from its representation. The intuition is that if a network can accurately restore the original image from a condensed representation, such representations should be informative for a variety of downstream tasks. In \cite{haghighi2021transferable}, the authors use a combination of patient-level contrastive learning—predicting a patient identifier from a patch-level representation, and a patch-level restorative approach—predicting occluded regions of the image. Meanwhile, \cite{tang2022self} contrasts and restores patches of the same image using a Swin UNETR \cite{hatamizadeh2021swin}.

\subsection{Diffusion as a \textit{Pretext} Task}
Recently, diffusion models have emerged as a leading method for image generation \cite{ho2020denoising}. The diffusion process begins with a sample \( x_T \) drawn from a standard Gaussian distribution \( x_T \sim \mathcal{N}(0, I) \). This sample is progressively denoised through a neural network that approximates the reverse Markov process:

$$
p_{\theta}(x_{t-1}|x_t) \sim \mathcal{N}(\mu_{\theta}(x_t, t), \Sigma_{\theta}(x_t, t)),
$$ where \( \mu_{\theta}(x_t, t) \) represents a linear combination of \( x_t \) and the predicted noise \( \epsilon_{\theta}(x_t, t) \) by the noise prediction network. Typically, U-shaped networks, such as UNet or Feature Pyramid Network, are employed for \( \epsilon_{\theta}(x_t, t) \) to model the noise.

In \cite{baranchuk2021label}, the authors introduce an innovative approach to semantic segmentation in natural images by leveraging the diffusion process as a pretraining method. The core idea involves applying Gaussian noise to images and iteratively restoring them using an FPN architecture. This process generates hierarchical feature maps at each denoising step. These maps are then utilized as frozen feature representations to train an ensemble of multi-layer perceptrons for the downstream task of semantic segmentation.

\section{Methods}
\label{sec:pagestyle}

% In this part we will give a brief overview of diffusion models. Next, we will describe how we use them for the pretrain problem.  At the end we will show how we adapted them for medical data.

\subsection{Pretrain}
As a pretrein task, we use the image generation task, but due to the specific domain, several problems arise. The original medical images have too high resolution, for certainty , we denote them $H, W, D$, so we divided them into patches, i.e. we taught diffusion models to generate patches of some specific size $h, w, d.$ On the inference , a picture of size $H, W, D$ will be divided into patches of size $h, w, d$ with some overlapping factor. Inside the intersection activations will be aggregated by taking mean values.

\subsection{Adaptive pretrain}
\label{sec:pretrain}
Due to the problem that arises due to certain domain specifics, we decided to modify our model to capture additional information. The hypothesis is that because of this separation of the picture, it is difficult for the model to restore geometric relationships in the picture. It has been shown that diffusion models are well conditioned on some additional information. As a result, we decided to throw in additional information about the physical coordinates in the picture. This should improve the ability to generate pictures by capturing the geometry of the picture and, as a result, improve the result on the final task. We can take the coordinates from the pretrained body part regressor (BPR) model. The BPR model takes an image as input and outputs a natural geometric coordinate for each voxel in the image. It was trained based on ~\cite{bpr}. In fig.~\ref{fig:pipeline} this step is depicted by \textit{concatenate channel-wise} procedure.

\subsection{Network architecture}
% \todo{Write about FPN and BPR, include information on the number of FLOPs}
We have tested a different number of architectures. Initially, due to the specifics of the data domain, we tried not to use architectures with an attention structure. The architecture was made purely from convolutions. In further experiments, we found that the capacity of such models is not enough to capture small details of the picture. We decided to use an adapted attention architecture linear in the input of the image~\cite{rui}. It is used at all levels of Unet except the last ones, where the number of channels begins to greatly exceed the size of the feature maps, so in them we used a standard quadratic attention blocks in size of the features, but linear in channel space. This greatly improved the quality of training. Later we used this architecture.
% \begin{figure}[htb]

% \begin{minipage}[b]{1.0\linewidth}
%   \centering
%   \centerline{\includegraphics[width=9cm]{images/attention.png}}
% %  \vspace{2.0cm}
%  \caption{MAE metric during training procedure for two Diffusion models with attention blocks and without}
%  \label{fig:attention}
% \end{minipage}

% \end{figure}

\subsection{Segmentation with pretrain model}
After training the model on the auxiliary generation task, we use it for the segmentation task. The whole process is shown in fig.~\ref{fig:pipeline}. 
We initially choose timesteps $t \in \{0, \dots, T\}$ based on our experiments. After that, the algorithm consists in that from the initial image get $x_{t}$ applying the formula: $$q(x_t| x_0) = \mathcal{N}(x_t| \sqrt{\overline{\alpha}_t} x_0, (1 - \overline{\alpha}_t)I).$$
After that, we can get embedding from each noisy image by applying a trained model $\epsilon_{\theta} (x_t, t)$ for each image. The model can be conditioned to coordinates as mentioned in par.~\ref{sec:pretrain}. After that we use feature maps from the model outputs for semantic segmentation. The final segmentation occurs due to nonlinear probing.

\begin{figure}[htb]

\begin{minipage}[b]{1.0\linewidth}
  \centering
  \centerline{\includegraphics[width=9cm]{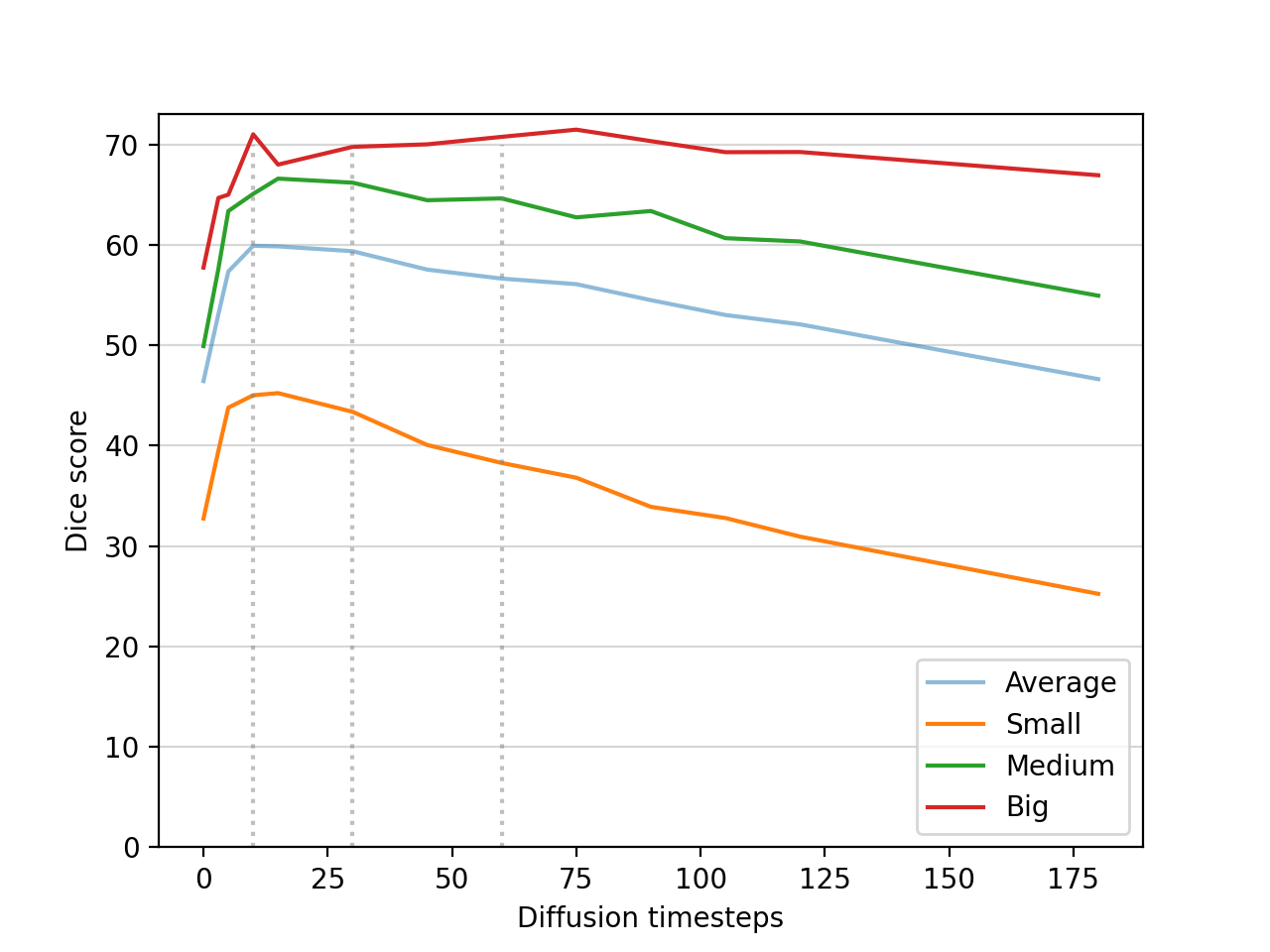}}
%  \vspace{2.0cm}
 \caption{Ablation of the number of diffusion timesteps. The results of image segmentation for \underline{objects of different sizes}: Big, Medium, and Small, non-linear probing regime. Average corresponds to all organs' average. Recall how segmentation quality spikes around 10 diffusion steps and then degrades for Medium and Small objects, while plateauing for Big objects.}
 \label{fig:2}
\end{minipage}

\end{figure}

\begin{table*}[]
\begin{center}
\begin{tabular}{ccccccccccccc}
    \toprule
    model & Sp   & Kid  & Gb   & Es   & Li   & St   & Aor  & IVC  & PSV  & Pa & AG & Avg  \\
    \midrule
    TransVW      & $24.9$ & $31.5$ & $6.7 $ & $28.1$  & $45.1 $ & $9.0$   & $44.9$ & $27.2$ & $19.0$ & $7.2 $ & $15.4$ &  $23.5$\\
    random-FPN   & $76.7$ & $67.0$ & $34.1$ &$ 47.1$ &$ 83.7$ &$ 52.8$ & $70.2$ & $57.5$ & $30.2$ & $28.6$ &$ 31.5$ &$ 52.1$ \\
    SwinUNETR    & $77.0$ & $74.4$ & $48.1$ &$ 52.1$ &$ 87.0$ &$ 53.7$ & $73.5$ & $58.1$ & $47.2$ & $35.3$ &$ 39.9$ &$ 58.5$ \\
    vox2vec-UNet & $80.3$ & $81.4$ & $34.1$ &$ 42.7$ &$ 91.1$ &$ 64.0$ & $79.6$ & $71.6$ & $42.7$ & $43.3$ &$ 37.6$ &$ 60.6$ \\
    vox2vec-FPN  & \textbf{91.0} & \textbf{89.2} & \textbf{50.7} & \underline{$67.5$} & \textbf{95.3} &\textbf{78.2} & \textbf{89.4} & \textbf{80.7} & \textbf{64.9} & \textbf{66.1} &\textbf{59.9} &\textbf{75.5} \\
    DDPM-FPN&    \underline{$82.3$} & \underline{$85.3$} & \underline{$37.1$} & \textbf{68.1} &	$87.8$&	\underline{$64.8$} &$83.9$  & $73.8$ &	\underline{$49.0$}&	$44.4$ &$48.4$  & $66.0$\\
    $\text{DDPM}_{\text{BPR}}$-FPN & $82.3$	& $85.7$	& $34.1$	& $62.0$	& \underline{$89.1$}	& $60.0$	& \underline{$86.1$}	& \underline{$74.0$}	& $48.5$	& \underline{$53.9$}	& \underline{$49.1$}	& \underline{$67.8$} \\
    \bottomrule
    
\label{tab:btcv}
\end{tabular}
\caption{Segmentation performance on BTCV dataset in non-linear probing regime on top of frozen representations. Best results are in bold, second best are underlined.}
% \todo{replace results for BPR+Diffusion} }
\end{center}
\end{table*}

\begin{table*}[]
\begin{center}
\begin{tabular}{ccccccccccccc}
    \toprule
    model & Small   & Medium  &  Big & AVG  \\
    \midrule
    random-FPN   & $48.9$ & $66.1$ & $78.3$ &$ 63.4$  \\
    SwinUNETR    & $64.8$ & $76.5$ & $87.6$ &$ 75.4$  \\
    vox2vec-UNet & $59.2$ & $73.4$ & $86.4$ &$ 71.9$ \\
    vox2vec-FPN  & \textbf{77.9} & \textbf{84.0} & \textbf{92.5} &\textbf{ 84.1}  \\
    DDPM-FPN&    \underline{$67.9$}  & \underline{$78.6$}& \underline{$88.6$} & \underline{$77.5$} \\
    $\text{DDPM}_{\text{BPR}}$-FPN & $66.5$	& $77.1$	& $85.1$	& $75.6$	 \\
    \bottomrule
    
\label{tab:flare}
\end{tabular}
\caption{Segmentation performance on FLARE dataset in non-linear probing regime on top of frozen representations and probing head. Best results are in bold, second best are underlined.}
\end{center}
\end{table*}

\subsection{Data}
We use our method to pre-train the FPN architecture on six public computed tomography datasets ~\cite{amos,flare,nlst,nsclc,lidc}. In total, we utilize over 6500 CT scans, covering thoracic and chest regions. Although some of these datasets come with dense segmentation annotations, we do not use them during the pre-training stage. Both the diffusion model and the body-part regressor models use the same pretraining data and are trained separately. 

% \todo{Data preprocessing, data augmentations, how many epochs, lr, optimizer, batch size, augmentations, GPU details}

We scale all images to one spacing equal to (1, 1, 2) and divided them into patches during the training into sizes (128, 128, 32). Each model was trained for 3000 epochs with learning rate is equal to $10^{-4}$ and an Adam optimizer. We used one A100 GPU card with 40 GB of memory and due to lack of resources, we trained the model with batch size equal to 1.

% \todo{Modified}

To test the effectiveness of our method, we use two datasets with a multi-class organ segmentation task. We train models on the Beyond the Cranial Vault Abdomen. \textbf{BTCV} \cite{btcv} consists of 30 CT scans with annotations for 13 different organs, namely: spleen (Sp), right and left kidneys (Kid), gallbladder (Gb), esophagus (Es), liver (Li), stomach (St), aorta (Aor), inferior vena cava (IVC), portal and spleen veins (PSV), pancreas (Pa), and right and left adrenal glands (AG). All augmentations with images are the same as in pretrain procedure. We freezed diffusion weights in this step and trained only probing head. For an honest comparison, we used all hyperparameters similar to ~\cite{goncharov2023vox2vec} for all models.

% \todo{Modified}

Additionally, we apply trained models on the Fast and Low-resource semi-supervised Abdominal organ segmentation in CT dataset \cite{flare}, \textbf{FLARE}. It includes 50 CT scans with organ segmentations. A total of 12 organ annotations coincide between the two datasets, with portal and spleen veins (PSV) missing from FLARE.  To compare models' capabilities to segment objects of different sizes we group organs into three categories. Big: spleen, liver, stomach; medium: kidneys, esophagus, aorta, pancreas; and small: gallbladder, inferior vena cava, portal and spleen veins  and adrenal glands.

\section{Results}
%  SwinUNETR and TransVW cost 391 GFLOPs
% and 1.2 TFLOPS, correspondingly, compared to 115 GFLOPs of vox2vec-FPN
% \todo{
% \begin{enumerate}
%     \item Detailed analysis depending on the size of the ROI (Figure + table)
%     \item Reconstraction curves (Figure, re-draw)
%     \item Fig. 5 re-draw, same pic? 
%     \item (Consider)  Think of a way to demonstrate the effect of BPR
%     \item (Consider) FLOPs vs Dice on Big, Medium, Small
%     \item (Consider) RUN inference (only inference) of models on FLARE or AMOS, to demonstrate the DROP in performance. Hopefully the drop of our models will be less than the baselines.
% \end{enumerate}}

\subsection{Timestemps evaluation}
We investigated which timestamps are best suited for our task. We evaluated our pre-trained model for the segmentation problem of 13 classes for the BTCV dataset with nonlinear probing. We divided the classes into small, medium and large in size to analyze how well the model takes the semantics of data of various sizes. Average results have shown in Fig.~\ref{fig:2}.

To explain the results of the experiment, we consider the problem of generating an image using a diffusion model as a solution by the model of a huge number of denoising tasks $x_t \rightarrow x_0$. Note that too huge timestep show a bad result, because the picture is too noisy and it's hard to get representative features out of it. On the other hand, too small timestep worsen the result because the task becomes too simple and the model produces not very representative features. For each dataset, you need to find your best $t$ at which the quality will be the best. As a universal approach, we suggest taking 3 different timesteps and using them together. Based on results of experiments, we selected 3 different timesteps $t_1 = 10, t_2 = 30,t_3 = 60.$

% \label{sec:typestyle}

% \begin{figure}[htb]

% \begin{minipage}[b]{.48\linewidth}
%   \centering
%   \centerline{\includegraphics[width=4.0cm]{images/2.jpg}}
% %  \vspace{1.5cm}
%   \centerline{}\medskip
% \end{minipage}
% \hfill
% \begin{minipage}[b]{0.48\linewidth}
%   \centering
%   \centerline{\includegraphics[width=4.0cm]{images/3.jpg}}
% %  \vspace{1.5cm}
%   \centerline{}\medskip
% \end{minipage}
% %
% \begin{minipage}[b]{.48\linewidth}
%   \centering
%   \centerline{\includegraphics[width=4.0cm]{images/4.jpg}}
% %  \vspace{1.5cm}
%   \centerline{}\medskip
% \end{minipage}
% \hfill
% \begin{minipage}[b]{0.48\linewidth}
%   \centering
%   \centerline{\includegraphics[width=4.0cm]{images/5.jpg}}
% %  \vspace{1.5cm}
%   \centerline{}\medskip
% \end{minipage}
% %
% \caption{Example of the same coordinate from BPR for 4 different CT images.}
% \label{fig:3}
% %
% \end{figure}

% To evaluate the effectiveness of the model on a proxy task, we consider a reconstruction metric. In our case we choose have chosen MAE between the restored picture and the original one. The fig.~\ref{fig:4} shows the metric on the validation part of dataset during training. You can see that in average the conditioned model converges faster and in the end has better metric score. The fig.~\ref{fig:5} also shows reconstructions for some slice of CT image.
\begin{figure}[htb]

% \begin{minipage}[b]{1.0\linewidth}
%   \centering
%   \centerline{\includegraphics[width=9cm]{images/6.png}}
% %  \vspace{2.0cm}
%  \caption{Reconstruction loss with BPR and without.}
%  \label{fig:4}
% \end{minipage}

\end{figure}

% \begin{figure}[htb]

% \begin{minipage}[b]{1.0\linewidth}
%   \centering
%   \centerline{\includegraphics[width=8.5cm]{images/8.jpg}}
% %  \vspace{2.0cm}
%   \centerline{Diffusion}\medskip
% \end{minipage}
% %
% \begin{minipage}[b]{1.0\linewidth}
%   \centering
%   \centerline{\includegraphics[width=8.5cm]{images/7.jpg}}
% %  \vspace{2.0cm}
%   \centerline{Diffusion with BPR}\medskip
% \end{minipage}
% %
% \caption{The original picture is on the left, noisy in the center, and on the right reconstructed image for different pretrained model.}
% \label{fig:5}
% %
% \end{figure}

\subsection{Probing}
% We would like to remind you that our work is not a SOTA method and we do not seek to beat existing methods. Our work is proof of concept that this methodology really works and it can and should be improved. 

We analyzed our models on BTCV dataset in a non-linear probing regime, which is a small convolutional head on top of frozen (after pretraining) FPN features. The results are shown in the Tables 1, 2. Models trained on BTCV were additionally evaluated on the FLARE dataset, which provide similar segmentation classes. All models demonstrate an increased performance on this transfer learning experiment. 
% \ref{tab:btcv}, \ref{tab:flare}. 
Our model outperforms both restorative pretraining methods SwinUnetr and TransVW, being second only to the state-of-the-art contrastive pretrain method vox2vec.

% \subsubsection{BTCV}

% % Below is an example of how to insert images. Delete the ``\vspace'' line,
% % uncomment the preceding line ``\centerline...'' and replace ``imageX.ps''
% % with a suitable PostScript file name.
% % -------------------------------------------------------------------------
% \begin{figure}[htb]

% \begin{minipage}[b]{1.0\linewidth}
%   \centering
%   \centerline{\includegraphics[width=8.5cm]{example-image}}
% %  \vspace{2.0cm}
%   \centerline{(a) Result 1}\medskip
% \end{minipage}
% %
% \begin{minipage}[b]{.48\linewidth}
%   \centering
%   \centerline{\includegraphics[width=4.0cm]{example-image}}
% %  \vspace{1.5cm}
%   \centerline{(b) Results 3}\medskip
% \end{minipage}
% \hfill
% \begin{minipage}[b]{0.48\linewidth}
%   \centering
%   \centerline{\includegraphics[width=4.0cm]{example-image}}
% %  \vspace{1.5cm}
%   \centerline{(c) Result 4}\medskip
% \end{minipage}
% %
% \caption{Example of placing a figure with experimental results.}
% \label{fig:res}
% %
% \end{figure}

% % To start a new column (but not a new page) and help balance the last-page
% % column length use \vfill\pagebreak.
% % -------------------------------------------------------------------------
% \vfill
% \pagebreak

% References should be produced using the bibtex program from suitable

\section{Conclusion}

In conclusion, this study introduces an innovative pretraining strategy utilizing diffusion models guided by anatomical insights for enhancing 3D medical image segmentation. By integrating an auxiliary diffusion process and a model predicting universal 3D body-part coordinates, our approach effectively generates generalizable feature representations, crucial for various downstream segmentation tasks. This method not only overcomes the limitations of existing restorative pretraining methods but also competes closely with leading contrastive pretraining approaches.
\section{Compliance with Ethical Standards}
This research study was conducted retrospectively using human subject data made available in open access by \cite{btcv,flare}. Ethical approval was not required as confirmed by the license attached with the open access data.

\section{Acknowledgments}
No funding was received for conducting this study. The authors have no relevant financial or non-financial interests to disclose.
\bibliographystyle{IEEEbib}
\bibliography{strings,refs}

\begin{thebibliography}{10}

\bibitem{he2022masked}
Kaiming He, Xinlei Chen, Saining Xie, Yanghao Li, Piotr Doll{\'a}r, and Ross Girshick,
\newblock ``Masked autoencoders are scalable vision learners,''
\newblock in {\em Proceedings of the IEEE/CVF conference on computer vision and pattern recognition}, 2022, pp. 16000--16009.

\bibitem{taleb20203d}
Aiham Taleb, Winfried Loetzsch, Noel Danz, Julius Severin, Thomas Gaertner, Benjamin Bergner, and Christoph Lippert,
\newblock ``3d self-supervised methods for medical imaging,''
\newblock {\em Advances in neural information processing systems}, vol. 33, pp. 18158--18172, 2020.

\bibitem{haghighi2021transferable}
Fatemeh Haghighi, Mohammad Reza~Hosseinzadeh Taher, Zongwei Zhou, Michael~B Gotway, and Jianming Liang,
\newblock ``Transferable visual words: Exploiting the semantics of anatomical patterns for self-supervised learning,''
\newblock {\em IEEE transactions on medical imaging}, vol. 40, no. 10, pp. 2857--2868, 2021.

\bibitem{tang2022self}
Yucheng Tang, Dong Yang, Wenqi Li, Holger~R Roth, Bennett Landman, Daguang Xu, Vishwesh Nath, and Ali Hatamizadeh,
\newblock ``Self-supervised pre-training of swin transformers for 3d medical image analysis,''
\newblock in {\em Proceedings of the IEEE/CVF Conference on Computer Vision and Pattern Recognition}, 2022, pp. 20730--20740.

\bibitem{goncharov2023vox2vec}
Mikhail Goncharov, Vera Soboleva, Anvar Kurmukov, Maxim Pisov, and Mikhail Belyaev,
\newblock ``vox2vec: A framework for self-supervised contrastive learning of voxel-level representations in medical images,''
\newblock in {\em International Conference on Medical Image Computing and Computer-Assisted Intervention}. Springer, 2023, pp. 605--614.

\bibitem{hatamizadeh2021swin}
Ali Hatamizadeh, Vishwesh Nath, Yucheng Tang, Dong Yang, Holger~R Roth, and Daguang Xu,
\newblock ``Swin unetr: Swin transformers for semantic segmentation of brain tumors in mri images,''
\newblock in {\em International MICCAI Brainlesion Workshop}. Springer, 2021, pp. 272--284.

\bibitem{ho2020denoising}
Jonathan Ho, Ajay Jain, and Pieter Abbeel,
\newblock ``Denoising diffusion probabilistic models,''
\newblock {\em Advances in neural information processing systems}, vol. 33, pp. 6840--6851, 2020.

\bibitem{baranchuk2021label}
Dmitry Baranchuk, Ivan Rubachev, Andrey Voynov, Valentin Khrulkov, and Artem Babenko,
\newblock ``Label-efficient semantic segmentation with diffusion models,''
\newblock {\em arXiv preprint arXiv:2112.03126}, 2021.

\bibitem{bpr}
Ronald M.~Summers Ke~Yan, Le~Lu,
\newblock ``Unsupervised body part regression via spatially self-ordering convolutional neural networks,''
\newblock .

\bibitem{rui}
Chenxi~Duan Rui~Li, Jianlin~Su and Shunyi Zheng,
\newblock ``Linear attention mechanism: An efficient attention for semantic segmentation,''
\newblock .

\bibitem{amos}
Yuanfeng Ji, Haotian Bai, Chongjian Ge, Jie Yang, Ye~Zhu, Ruimao Zhang, Zhen Li, Lingyan Zhanng, Wanling Ma, Xiang Wan, et~al.,
\newblock ``Amos: A large-scale abdominal multi-organ benchmark for versatile medical image segmentation,''
\newblock {\em Advances in Neural Information Processing Systems}, vol. 35, pp. 36722--36732, 2022.

\bibitem{flare}
Jun Ma, Yao Zhang, Song Gu, Xingle An, Zhihe Wang, Cheng Ge, Congcong Wang, Fan Zhang, Yu~Wang, Yinan Xu, et~al.,
\newblock ``Fast and low-gpu-memory abdomen ct organ segmentation: the flare challenge,''
\newblock {\em Medical Image Analysis}, vol. 82, pp. 102616, 2022.

\bibitem{nlst}
``Data from the national lung screening trial (nlst),'' 2013.

\bibitem{nsclc}
HJWL Aerts, E~Rios Velazquez, RT~Leijenaar, Chintan Parmar, Patrick Grossmann, S~Cavalho, Johan Bussink, Ren{\'e} Monshouwer, Benjamin Haibe-Kains, Derek Rietveld, et~al.,
\newblock ``Data from nsclc-radiomics,''
\newblock {\em The cancer imaging archive}, 2015.

\bibitem{lidc}
Samuel~G Armato~III, Geoffrey McLennan, Luc Bidaut, Michael~F McNitt-Gray, Charles~R Meyer, Anthony~P Reeves, Binsheng Zhao, Denise~R Aberle, Claudia~I Henschke, Eric~A Hoffman, et~al.,
\newblock ``The lung image database consortium (lidc) and image database resource initiative (idri): a completed reference database of lung nodules on ct scans,''
\newblock {\em Medical physics}, vol. 38, no. 2, pp. 915--931, 2011.

\bibitem{btcv}
Bennett Landman, Zhoubing Xu, J~Igelsias, Martin Styner, T~Langerak, and Arno Klein,
\newblock ``Miccai multi-atlas labeling beyond the cranial vault--workshop and challenge,''
\newblock in {\em Proc. MICCAI Multi-Atlas Labeling Beyond Cranial Vault—Workshop Challenge}, 2015, vol.~5, p.~12.

\end{thebibliography}
\end{document}